%% file: report.tex
\documentclass{article} 
\usepackage{nips13submit_e,times}
\usepackage{graphicx}
\usepackage{amsmath,amssymb} 
\usepackage{multirow}
\usepackage{amsmath}
\usepackage{amssymb}
\usepackage{forloop}
\usepackage{calc}
\usepackage{rotating}
\usepackage[tight]{bibhacks}
\usepackage{enumitem}
\usepackage{xspace}
\usepackage{pgfplots}
\usepackage{tikz}
\usepackage{arydshln}

\usepackage{tikz}
\usetikzlibrary{arrows}

\title{Zero-Shot Learning by Convex Combination of Semantic Embeddings}

\author{Mohammad Norouzi\thanks{Part of this work was done while Mohammad Norouzi was at Google.}~,\, Tomas Mikolov,\, Samy Bengio,\, Yoram Singer,\\
{\bf Jonathon Shlens,\, Andrea Frome,\, Greg S. Corrado,\, Jeffrey Dean}\\[.2cm]
\texttt{norouzi@cs.toronto.edu,~\{tmikolov,~bengio,~singer\}@google.com}\\
\texttt{\{shlens,~afrome,~gcorrado,~jeff\}@google.com}\\[.2cm]
\begin{tabular}{c@{\hspace*{1.5cm}}c}
$^*$University of Toronto & Google, Inc.\\
ON, Canada & Mountain View, CA, USA\\
\end{tabular}
}

\input defs.tex

\nipsfinalcopy 

\begin{document}

\maketitle

\newcommand{\softmaxpp}{ConSE\xspace}
\newcommand{\n}{n}
\newcommand{\nway}{$\n$-way\xspace}
\newcommand{\eqdef}{\stackrel{\rm def}{=}}


\input{abstract}
\input{intro}
\input{model}
\input{results}
\input{conclusion}

\bibliographystyle{abbrv}
\small{
  \bibliography{myrefs}
}

\end{document}

%% file: defs.tex
\usepackage{amsfonts}
\usepackage{amsmath}
\usepackage{relsize}
\usepackage{algorithmic}

\newcommand{\x}{\vec{x}}

\newcommand{\s}{\vec{s}}

\def\by[#1]#2{#1~\hspace*{-.15cm}~\times~\hspace*{-.15cm}~#2}

\newcommand{\Dzset}{\mathcal D_0}
\newcommand{\Doset}{\mathcal D_1}

\newcommand{\Sset}{\mathcal S}
\newcommand{\Xset}{\mathcal X}

\newcommand{\Yzset}{\mathcal Y_0}
\newcommand{\Yoset}{\mathcal Y_1}

\renewcommand{\vec}[1]{\operatorname{vec}}
\let\originaleqref\eqref
\renewcommand{\eqref}{Eq.~\originaleqref}

\def\th#1{#1^\text{th}}

\newsavebox{\savepar}

\def\vec#1{{\bf #1}}

\def\Real{\mathbb{R}}

\newcommand{\tabref}[1]{Table~\ref{#1}}

\newcommand{\figref}[1]{Fig.~\ref{#1}}

\newcommand{\argmax}[1]{\underset{#1}{\operatorname{argmax}}}

\def\ie{{\em i.e.,}}

\def\eg{{\em e.g.,}}

\newcommand{\comment}[1]{}

%% file: abstract.tex
\begin{abstract}

Several recent publications have proposed methods for mapping images
into continuous semantic embedding spaces.  In some cases the
embedding space is trained jointly with the image transformation. In
other cases the semantic embedding space is established by an
independent natural language processing task, and then the image
transformation into that space is learned in a second stage.
Proponents of these image embedding systems have stressed their
advantages over the traditional \nway{} classification framing of
image understanding, particularly in terms of the promise for
zero-shot learning -- the ability to correctly annotate images of
previously unseen object categories.  In this paper, we propose a
simple method for constructing an image embedding system from any
existing \nway{} image classifier and a semantic word embedding model,
which contains the $\n$ class labels in its vocabulary.  Our method
maps images into the semantic embedding space via convex combination
of the class label embedding vectors, and requires no additional
training.  We show that this simple and direct method confers many of
the advantages associated with more complex image embedding schemes,
and indeed outperforms state of the art methods on the ImageNet
zero-shot learning task.

\end{abstract}

%% file: intro.tex
\section{Introduction}



The classic machine learning approach to object recognition
presupposes the existence of a large {\em labeled} training dataset to
optimize the free parameters of an image classifier.  There have been
continued efforts in collecting larger image corpora with a broader
coverage of object categories~(\eg~\cite{imagenet}), thereby enabling
image classification with many classes. While annotating more object
categories in images can lead to a finer granularity of image
classification, creating high quality fine grained image annotations
is challenging, expensive, and time consuming. Moreover, as new visual
entities emerge over time, the annotations should be revised, and the
classifiers should be re-trained.

Motivated by the challenges facing standard machine learning framework
for \nway{} classification, especially when $\n$ (the number of class
labels) is large, several recent papers have proposed methods for
mapping images into semantic embedding
spaces~\cite{Palatucci09,Farhadi09,Lampert09,devise,Socher13,
Weston2011wsabie}. In doing so, it is hoped that by resorting to
nearest neighbor search in the embedding space with respect to a set
of label embedding vectors, one can address {\em zero-shot learning}
-- annotation of images with new labels corresponding to previously
unseen object categories. While the common practice for image
embedding is to learn a regression model from images into a semantic
embedding space, it has been unclear whether there exists a more
direct way to transform any probabilistic \nway{} classifier into an
image embedding model, which can be used for zero-shot learning. In
this work, we present a simple method for constructing an image
embedding system by combining any existing probabilistic \nway{} image
classifier with an existing word embedding model, which contains the
$\n$ class labels in its vocabulary. We show that our simple method
confers many of the advantages associated with more complex image
embedding schemes.


Recently, zero-shot learning~\cite{larochelle2008zero,Palatucci09} has
received a growing amount of
attention~\cite{Rohrbach11,mensink2012metric,devise,Socher13}. A key
to zero-shot learning is the use of a set of semantic embedding
vectors associated with the class labels. These semantic embedding
vectors might be obtained from human-labeled object
attributes~\cite{Farhadi09,Lampert09}, or they might be learned from a
text corpus in an unsupervised fashion, based on an independent
natural language modeling
task~\cite{devise,Socher13,skipgram}. Regardless of the way the label
embedding vectors are obtained, previous work casts zero-shot learning
as a regression problem from the input space into the embedding
space. In contrast, given a pre-trained standard classifier, our
method maps images into the semantic embedding space via the convex
combination of the class label embedding vectors. The values of a
given classifier's predictive probabilities for different training
labels are used to compute a weighted combination of the label
embeddings in the semantic space. This provides a continuous embedding
vector for each image, which is then used for extrapolating the
pre-trained classifier's predictions beyond the {\em training} labels,
into a set of {\em test} labels.

The effectiveness of our method called ``convex combination of
semantic embeddings'' (\softmaxpp{}) is evaluated on ImageNet
zero-shot learning task.  By employing a convolutional neural
network~\cite{Krizhevsky12} trained only on $1000$ object categories
from ImageNet, the \softmaxpp{} model is able to achieve $9.4\%$
hit@$1$ and $24.7\%$ hit@$5$ on $1600$ unseen objects categories,
which were omitted from the training dataset. When the number of test
classes gets larger, and they get further from the training classes in
the ImageNet category hierarchy, the zero-shot classification results
get worse, as expected, but still our model outperforms a recent
state-of-the-art model~\cite{devise} applied to the same task.


%% file: model.tex
\section{Previous work}

Zero-shot learning is closely related to {\em one-shot
  learning}~\cite{miller2000, fei2006one, bart2005cross, lake2011one},
where the goal is to learn object classifiers based on a few labeled
training exemplars. The key difference in zero-shot learning is that
no training images are provided for a held-out set of test
categories. Thus, zero-shot learning is more challenging, and the use
of side information about the interactions between the class labels is
more essential in this setting. Nevertheless, we expect that advances
in zero-shot learning will benefit one-shot learning, and visual
recognition in general, by providing better ways to incorporate prior
knowledge about the relationships between the object categories.

A key component of zero-shot learning is the way a semantic space of
class label embeddings is defined. In computer vision, there has been
a body of work on the use of human-labeled visual
attributes~\cite{Farhadi09,Lampert09} to help detecting unseen object
categories. {\em Binary} attributes are most commonly used to encode
presence and absence of a set of visual characteristics within
instances of an object category. Some examples of these attributes
include different types of materials, different colors, textures, and
object parts. More recently, relative attributes~\cite{parikh2011} are
proposed to strengthen the attribute based representations. In
attribute based approaches, each class label is represented by a
vector of attributes, instead of the standard one-of-$\n$
encoding. And multiple classifiers are trained for predicting each
object attribute. While this is closely related to our approach, the
main issue with attribute-based classification is its lack of
scalability to large-scale tasks. Annotating thousands of attributes
for thousands of object classes is an ambiguous and challenging task
in itself, and the applicability of supervised attributes to
large-scale zero-shot learning is limited. There has been some recent
work showing good zero-shot classification performance on visual
recognition tasks~\cite{rohrbach2011evaluating, mensink2012metric},
but these methods also rely on the use of knowledge bases containing
descriptive properties of object classes, and the WordNet hierarchy.

A more scalable approach to semantic embeddings of class labels builds
upon the recent advances in unsupervised neural language
modeling~\cite{bengio06}. In this approach, a set of multi-dimensional
embedding vectors are learned for each word in a text corpus. The word
embeddings are optimized to increase the predictability of each word
given its context~\cite{skipgram}. Essentially, the words that
cooccur in similar contexts, are mapped onto similar embedding
vectors. Frome et al.~\cite{devise} and Socher et al.~\cite{Socher13}
exploit such word embeddings to embed textual names of object class
labels into a continuous semantic space. In this work, we also use the
skip-gram model~\cite{skipgram} to learn the class label embeddings.

\section{Problem Statement}

Assume that a labeled training dataset of images $\Dzset \equiv \{
(\x_i,\, y_i) \}_{i=1}^m$ is given, where each image is represented by
a $p$-dimensional feature vector, $\x_i \in \Real^p$. For generality
we denote $\Xset \eqdef \Real^p$.
There are $\n_0$ distinct class labels available for
training, \ie~$y_i \in \Yzset \equiv \{1, \ldots, \n_0\}$. In addition,
a test dataset denoted $\Doset \equiv \{ (\x'_j,\, y'_j)
\}_{j=1}^{m'}$ is provided, where $\x'_j \in \Xset$ as above, while
$y'_j \in \Yoset \equiv \{\n_0\!+\!1, \ldots, \n_0\!+\!\n_1\}$. The test
set contains $\n_1$ distinct class labels, which are omitted from the
training set. Let $\n = \n_0 \!+\!  \n_1$ denote the total number of
labels in the training and test sets.

The goal of zero-shot learning is to train a classifier on the
training set $\Dzset$, which performs reasonably well on the unseen
test set $\Doset$. Clearly, without any side information about the
relationships between the labels in $\Yzset$ and $\Yoset$, zero-shot
learning is infeasible as $\Yzset \cap \Yoset =
\varnothing$. However, to mitigate zero-shot learning, one typically
assumes that each class label $y$ ($1 \le y \le \n$) is associated
with a semantic embedding vector $s(y) \in \Sset \equiv \Real^q$. The
semantic embedding vectors are such that two labels $y$ and $y'$ are
similar if and only if their semantic embeddings $s(y)$ and $s({y'})$
are close in~$\Sset$, \eg~$\langle s(y), s({y'}) \rangle_{\Sset}$ is
large. Clearly, given an embedding of training and test class labels
into a joint semantic space \ie~$\{ \s(y);~y \in \Yzset \cup \Yoset
\}$, the training and test labels become related, and one can hope to
learn from the training labels to predict the test labels.


Previous work (\eg~\cite{devise, Socher13}) has addressed zero-shot
classification by learning a mapping from input features to semantic
label embedding vectors using a multivariate regression
model. Accordingly, during training instead of learning an $\n_0$-way
classifier from inputs to training labels ($\Xset \to \Yzset$), a
regression model is learned from inputs to the semantic embedding
space ($\Xset \to \Sset$). A training dataset of inputs paired with
semantic embeddings, \ie~$\{(\x_i, s({y_i}));\,(\x_i,
y_i) \in \Dzset\}$, is constructed to train a regression function $f
: \Xset \to \Sset$ that aims to map $\x_i$ to $s({y_i})$. Once
$f(\cdot)$ is learned, it is applied to a test image $\x'_j$ to obtain
$f(\x'_j)$, and this continuous semantic embedding for $\x'_j$ is then
compared with the test label embedding vectors,
$\{s({y'});~y' \in \Yoset \}$, to find the most relevant test
labels. Thus, instead of directly mapping from $\Xset \to \Yoset$,
which seems impossible, zero-shot learning methods first learn a
mapping $\Xset \to \Sset$, and then a deterministic mapping such as
$k$-nearest neighbor search in the semantic space is used to map a
point in $\Sset$ to a ranked list of labels in $\Yoset$.

\section{\softmaxpp{}: Convex combination of semantic embeddings}

\subsection{Model Description}

In contrast to previous work which casts zero-shot learning as a
regression problem from the input space to the semantic label
embedding space, in this work, we do not explicitly learn a regression
function $f : \Xset \to \Sset$. Instead, we follow the classic machine
learning approach, and learn a classifier from training inputs to
training labels. To this end, a classifier $p_0$ is trained on
$\Dzset$ to estimate the probability of an image $\x$ belonging to a
class label $y \in \Yzset$, denoted $p_0(y \mid\x)$, where
$\sum_{y=1}^{\n_0} {p_0(y \mid \x)} = 1$. Given $p_0$, we propose a
method to transfer the probabilistic predictions of the classifier
beyond the training labels, to a set of test labels.

Let $\widehat{y}_0(\x, 1)$ denote the most likely training label for
an image $\x$ according to the classifier $p_0$. Formally, we denote
\begin{equation}
\widehat{y}_0(\x, 1) \equiv \argmax{y \in \Yzset}~p_0(y \mid \x)~.
\end{equation}
Analogously, let $\widehat{y}_0(\x, t)$ denote the $\th{t}$ most
likely training label for $\x$ according to $p_0$. In other words,
$p_0(\widehat{y}_0(\x, t)\mid\x)$ is the $\th{t}$ largest value among
$\{ p_0(y \mid \x);~y \in \Yzset \}$. Given the top $T$ predictions of
$p_0$ for an input $\x$, our model deterministically predicts a
semantic embedding vector $f(\x)$ for an input $\x$, as the convex
combination of the semantic embeddings $\{ s(\widehat{y}_0(\x, t))
\}_{t=1}^T$ weighted by their corresponding probabilities. More
formally,
\begin{equation}
f(\x) = \frac{1}{Z} \sum_{t=1}^T
  p(\widehat{y}_0(\x, t) \mid \x) \cdot s(\widehat{y}_0(\x, t)) ~ ,
\label{eq:embedding}
\end{equation}
where $Z$ is a normalization factor given by $Z = \sum_{t=1}^T
p(\widehat{y}_0(\x, t) \mid \x)$, and $T$ is a hyper-parameter
controlling the maximum number of embedding vectors to be considered.
If the classifier is very confident in its prediction of a label $y$
for $\x$, \ie~$p_0(y \mid \x) \approx 1$, then $f(\x) \approx
s(y)$. However, if the classifier have doubts whether an image
contains a ``lion'' or a ``tiger'', \eg~$p_0(\text{lion} \mid \x) =
0.6$ and $p_0(\text{tiger} \mid \x) = 0.4$, then our predicted
semantic embedding, $f(\x) = 0.6 \cdot s(\text{lion}) + 0.4 \cdot
s(\text{tiger})$, will be something between lion and tiger in the
semantic space.  Even though ``liger'' (a hybrid cross between a lion
and a tiger) might not be among the training labels, because it is
likely that $s(\text{liger}) \approx \frac{1}{2}s(\text{lion}) +
\frac{1}{2}s(\text{tiger})$, then it is likely that $f(\x) \approx
s(\text{liger})$.

Given the predicted embedding of $\x$ in the semantic space,
\ie~$f(\x)$, we perform zero-shot classification by finding the class
labels with embeddings nearest to $f(\x)$ in the semantic space. The
top prediction of our model for an image $\x$ from the test label set,
denoted $\widehat{y}_1(\x, 1)$, is given by
\begin{equation}
\widehat{y}_1(\x, 1) \equiv \argmax{y' \in \Yoset}~cos(f(\x), s(y'))~,
\label{eq:zeroshot-extrapolate}
\end{equation}
where we use cosine similarity to rank the embedding vectors.
Moreover, let $\widehat{y}_1(\x, k)$ denote the $\th{k}$ most likely
test label predicted for $\x$. Then, $\widehat{y}_1(\x, k)$ is defined
as the label $y' \in \Yoset$ with the $\th{k}$ largest value of cosine
similarity in $\{ cos(f(\x), s(y'));~y' \in \Yoset \}$. Note that
previous work on zero-shot learning also uses a similar $k$-nearest
neighbor procedure in the semantic space to perform label
extrapolation. The key difference in our work is that we define the
embedding prediction $f(\x)$ based on a standard classifier as in
\eqref{eq:embedding}, and not based on a learned regression model.
For the specific choice of cosine similarity to measure closeness in
the embedding space, the norm of $f(\x)$ does not matter, and we could
drop the normalization factor $(1/Z)$ in \eqref{eq:embedding}.

\subsection{Difference with DeViSE}

Our model is inspired by a technique recently proposed for image
embedding, called ``Deep Visual-Semantic Embedding''
(DeViSE)~\cite{devise}. Both DeViSE and \softmaxpp{} models benefit
from the convolutional neural network classifier of Krizhevsky et
al.~\cite{Krizhevsky12}, but there is an important difference in the
way they employ the neural net. The DeViSE model replaces the last
layer of the convolutional net, the Softmax layer, with a linear
transformation layer. The new transformation layer is trained using a
ranking objective to map training inputs close to continuous embedding
vectors corresponding to correct labels. Subsequently, the lower
layers of the convolutional neural network are fine-tuned using the
ranking objective to produce better results. In contrast, the
\softmaxpp{} model keeps the Softmax layer of the convolutional net
intact, and it does not train the neural network any further. Given a
test image, the \softmaxpp{} simply runs the convolutional classifier
and considers the top $T$ predictions of the model. Then, the convex
combination of the corresponding $T$ semantic embedding vectors in the
semantic space (see \eqref{eq:embedding}) is computed, which defines a
deterministic transformation from the outputs of the Softmax
classifier into the embedding space.

%% file: results.tex
\section{Experiments}

We compare our approach, ``convex combination of semantic embedding''
({\bf \softmaxpp{}}), with a state-of-the-art method called ``Deep
Visual-Semantic Embedding'' ({\bf DeViSE})~\cite{devise} on the
ImageNet dataset~\cite{imagenet}. Both of the \softmaxpp{} and DeViSE
models make use of the same skipgram text model~\cite{skipgram} to
define the semantic label embedding space.  The skipgram model was
trained on $5.4$ billion words from Wikipedia.org to construct $500$
dimensional word embedding vectors. The embedding vectors are then
normalized to have a unit norm. The convolutional neural network
of~\cite{Krizhevsky12}, used in both \softmaxpp{} and DeViSE, is
trained on ImageNet 2012 1K set with $1000$ training labels. Because
the image classifier, and the label embedding vectors are identical in
the \softmaxpp and DeViSE models, we can perform a direct comparison
between the two embedding techniques.

We mirror the ImageNet zero-shot learning experiments
of~\cite{devise}. Accordingly, we report the zero-shot generalization
performance of the models on three test datasets with increasing
degree of difficulty. The first test dataset, called ``2-hops''
includes labels from the 2011 21K set which are visually and
semantically similar to the training labels in the ImageNet 2012 1K
set. This dataset only includes labels within $2$ tree hops of the
ImageNet 2012 1K labels. A more difficult dataset including labels
within $3$ hops of the training labels is created in a similar way and
referred to as ``3-hops''. Finally, a dataset of all the labels in the
ImageNet 2011 21K set is created. The three test datasets respectively
include $1,589$, $7,860$, and $20,900$ labels.  These test datasets do
not include any image labeled with any of the $1000$ training labels.

\begin{figure}[t]
\begin{center}
\begin{tabular}{|@{\hspace{.05cm}}p{2.4cm}@{\hspace{.05cm}}|p{3.3cm}|p{3.3cm}|p{3.3cm}|}
\hline
&&&\\[-.25cm]
~Test Image & Softmax Baseline~\cite{Krizhevsky12} & DeViSE~\cite{devise} & \softmaxpp{\,($10$)} \\
\hline
&&&\\[-.35cm]
\includegraphics[width=2.4cm]{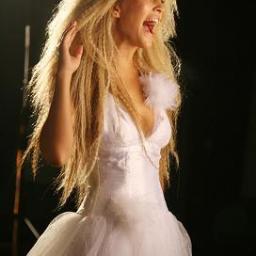} & 
\scriptsize
\vspace*{-2cm}
wig\newline
fur coat\newline
Saluki, gazelle hound\newline
Afghan hound, Afghan\newline
stole\newline
&
\scriptsize
\vspace*{-2cm}
water spaniel\newline
tea gown\newline
bridal gown, wedding gown\newline
spaniel\newline
tights, leotards
&
\scriptsize
\vspace*{-2cm}
business suit\newline
{\color{blue}\bf dress, frock}\newline
hairpiece, false hair, postiche\newline
swimsuit, swimwear, bathing suit\newline
kit, outfit
\\
\hdashline
&&&\\[-.35cm]
\includegraphics[width=2.4cm]{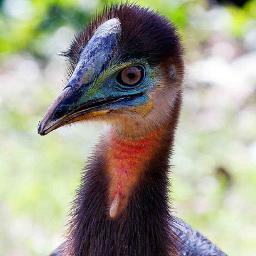} & 
\scriptsize
\vspace*{-2cm}
ostrich, Struthio camelus\newline
black stork, Ciconia nigra\newline
vulture\newline
crane\newline
peacock
& 
\scriptsize
\vspace*{-2cm}
heron\newline
owl,\,bird of Minerva,\,bird of night\newline
hawk\newline
bird of prey, raptor, raptorial bird\newline
finch
&
\scriptsize
\vspace*{-2cm}
{\color{blue}\bf ratite, ratite bird, flightless bird}\newline
peafowl, bird of Juno\newline
common spoonbill\newline
New World vulture, cathartid\newline
Greek partridge, rock partridge
\\
\hdashline
&&&\\[-.35cm]
\includegraphics[width=2.4cm]{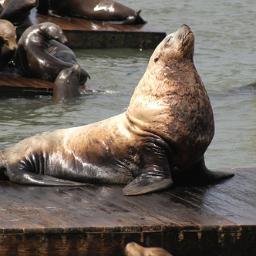} &
\scriptsize
\vspace*{-2cm}
sea lion\newline
plane, carpenter's plane\newline
cowboy boot\newline
loggerhead, loggerhead turtle\newline
goose
& 
\scriptsize
\vspace*{-2cm}
elephant\newline
turtle\newline
turtleneck, turtle, polo-neck\newline
flip-flop, thong\newline
handcart, pushcart, cart, go-cart\newline
&
\scriptsize
\vspace*{-2cm}
California sea lion\newline
{\color{blue}\bf Steller sea lion}\newline
Australian sea lion\newline
South American sea lion\newline
eared seal
\\
\hdashline
&&&\\[-.35cm]
\includegraphics[width=2.4cm]{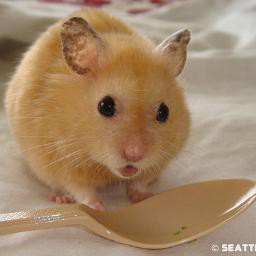} &
\scriptsize
\vspace*{-2cm}
hamster\newline
broccoli\newline
Pomeranian\newline
capuchin, ringtail
\newline
weasel &
\scriptsize
\vspace*{-2cm}
{\color{blue}\bf golden hamster, Syrian hamster}
\newline
rhesus, rhesus monkey
\newline
pipe\newline
shaker\newline
American mink, Mustela vison &
\scriptsize
\vspace*{-2cm}
{\color{blue}\bf golden hamster, Syrian hamster
}\newline
rodent, gnawer\newline
Eurasian hamster
\newline
rhesus, rhesus monkey
\newline
rabbit, coney, cony\\
\hdashline
&&&\\[-.35cm]
\includegraphics[width=2.4cm]{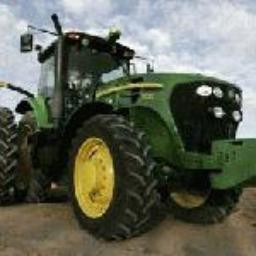} \newline
{\scriptsize \color{blue}\bf \hspace*{.4cm}(farm machine)} &
\scriptsize
\vspace*{-2cm}
thresher, 
threshing machine\newline
tractor\newline
harvester, reaper\newline
half track\newline
snowplow, snowplough
&
\scriptsize
\vspace*{-2cm}
truck, motortruck\newline
skidder\newline
tank car, tank\newline
automatic rifle, 
machine rifle\newline
trailer, house trailer
&
\scriptsize
\vspace*{-2cm}
flatcar, flatbed, flat\newline
truck, motortruck\newline
tracked vehicle\newline
bulldozer, dozer\newline
wheeled vehicle
\\
\hdashline
&&&\\[-.35cm]
\includegraphics[width=2.4cm]{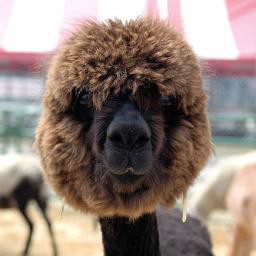} \vspace*{-.45cm} \newline
{\scriptsize \color{blue}\bf (alpaca, Lama pacos)} &
\scriptsize
\vspace*{-2cm}
Tibetan mastiff\newline
titi, titi monkey\newline
koala,\,koala bear,\,kangaroo bear
\newline
llama\newline
chow, chow chow
&
\scriptsize
\vspace*{-2cm}
kernel\newline
littoral, litoral, littoral zone, sands\newline
carillon\newline
Cabernet, Cabernet Sauvignon\newline
poodle, poodle dog
&
\scriptsize
\vspace*{-2cm}
dog, domestic dog
\newline
domestic cat, house cat
\newline
schnauzer\newline
Belgian sheepdog
\newline
domestic llama, Lama peruana
\\
\hline
\end{tabular}
\end{center}
\caption{Zero-shot test images from ImageNet, and their corresponding
  top 5 labels predicted by the Softmax Baseline~\cite{Krizhevsky12},
  DeViSE~\cite{devise}, and ConSE($T\!=\!10$). The labels predicted by the
  Softmax baseline are the labels used for training, and the labels
  predicted by the other two models are not seen during training of
  the image classifiers. The correct labels are shown in blue. Examples
  are hand-picked to illustrate the cases that the ConSE(10) performs
  well, and a few failure cases.}
\label{fig:qual-res}
\vspace*{-.2cm}
\end{figure}

\figref{fig:qual-res} depicts some qualitative results. The first
column shows the top $5$ predictions of the convolutional net,
referred to as the Softmax baseline~\cite{Krizhevsky12}. The second
and third columns show the zero-shot predictions by the DeViSE and
\softmaxpp{($10$)} models. The \softmaxpp{($10$)} model uses the top
$T = 10$ predictions of the Softmax baseline to generate convex
combination of embeddings. \figref{fig:qual-res} shows that the labels
predicted by the \softmaxpp{(10)} model are generally coherent and
they include very few outliers. In contrast, the top 5 labels
predicted by the DeViSE model include more outliers such as
``flip-flop'' predicted for a ``Steller sea lion'', ``pipe'' and
``shaker'' predicted for a ``hamster'', and ``automatic rifle''
predicted for a ``farm machine''.

The high level of annotation granularity in Imagenet, \eg~different
types of sea lions, creates challenges for recognition systems which
are based solely on visual cues. Using models such as \softmaxpp and
DeViSE, one can leverage the similarity between the class labels to
expand the original predictions of the image classifiers to a list of
similar labels, hence better retrieval rates can be achieved.

We report quantitative results in terms of two metrics: ``flat''
hit@$k$ and ``hierarchical'' precision@$k$. Flat hit@$k$ is the
percentage of test images for which the model returns the one true
label in its top $k$ predictions. Hierarchical precision@$k$ uses the
ImageNet category hierarchy to penalize the predictions that are
semantically far from the correct labels more than the predictions
that are close. Hierarchical precision@$k$ measures, on average, what
fraction of the model's top $k$ predictions are among the $k$ most
relevant labels for each test image, where the relevance of the labels
is measure by their distance in the Imagenet category hierarchy. A
more formal definition of hierarchical precision@$k$ is included in
the supplementary material of~\cite{devise}. Hierarchical
precision@$1$ is always equivalent to flat hit@$1$.

\begin{table}[t]
\begin{center}
\begin{tabular}{|l|c|l|c|c|c|c|c|}
\hline
~                                            & \# Candidate              & ~                    & \multicolumn{5}{c|}{Flat hit@$k$ (\%)} \\
\cline{4-8}
Test Label Set                               & Labels                    & Model                & 1     & 2     & 5     & 10    & 20    \\
\hline
\multirow{4}{*}{$2$-hops} & \multirow{4}{*}{$1,589$}  
                                                      & DeViSE                    & 6.0 &   10.0 &   18.1 &   26.4 &   36.4 \\
~                         & ~                         & \softmaxpp{(1)}           & 9.3 &   14.4 &   23.7 &   30.8 &   38.7 \\
~                         & ~                         & \softmaxpp{(10)}          & \bf 9.4 & \bf 15.1 & \bf 24.7 & \bf 32.7 & \bf 41.8 \\
~                         & ~                         & \softmaxpp{(1000)}        & 9.2 &   14.8 &   24.1 &   32.1 &   41.1 \\
\hdashline
\multirow{4}{*}{$2$-hops (+1K)} & ~  
                                        & DeViSE                    & \bf 0.8  &  2.7     &  7.9      & 14.2      & 22.7  \\
~                         & {$1,589$}   & \softmaxpp{(1)}           & 0.2 & \bf 7.1 & \bf 17.2 & 24.0 & 31.8 \\
~                         & {$+ 1000$}  & \softmaxpp{(10)}          & 0.3 & 6.2 & 17.0 & \bf 24.9 & \bf 33.5 \\
~                         & ~           & \softmaxpp{(1000)}        & 0.3 & 6.2 & 16.7 & 24.5 & 32.9 \\
\hline
\multirow{4}{*}{$3$-hops} & \multirow{4}{*}{$7,860$}  
                                                      & DeViSE                   & 1.7 & 2.9 & 5.3 & 8.2  & 12.5 \\
~                         & ~                         & \softmaxpp{(1)}           & 2.6	& 4.2 & 7.3 & 10.8 & 14.8 \\
~                         & ~                         & \softmaxpp{(10)}          & \bf 2.7 & \bf 4.4 & \bf 7.8 & \bf 11.5 & \bf 16.1 \\
~                         & ~                         & \softmaxpp{(1000)}        & 2.6	& 4.3 & 7.6 & 11.3 & 15.7 \\
\hdashline
\multirow{4}{*}{$3$-hops (+1K)} & ~  
                                        & DeViSE                    & \bf 0.5 & 1.4 & 3.4 & 5.9 & 9.7  \\
~                         & {$7,860$}   & \softmaxpp{(1)}           & 0.2 & \bf 2.4   & \bf 5.9 & 9.3 & 13.4 \\
~                         & {$+ 1000$}  & \softmaxpp{(10)}          & 0.2 & 2.2 & \bf 5.9 & \bf 9.7 & \bf 14.3 \\
~                         & ~           & \softmaxpp{(1000)}        & 0.2 & 2.2 & 5.8 & 9.5 & 14.0 \\
\hline
\multirow{4}{*}{ImageNet 2011 $21$K} & \multirow{4}{*}{$20,841$}  
                                                      & DeViSE                   & 0.8  & 1.4  & 2.5 & 3.9 & 6.0\\
~                         & ~                         & \softmaxpp{(1)}           & 1.3  & 2.1	& 3.6 & 5.4 & 7.6\\
~                         & ~                         & \softmaxpp{(10)}          & \bf 1.4  & \bf 2.2	& \bf 3.9 & \bf 5.8 & \bf 8.3\\
~                         & ~                         & \softmaxpp{(1000)}        & 1.3  & 2.1	& 3.8 & 5.6 & 8.1\\
\hdashline
\multirow{4}{*}{ImageNet 2011 $21$K (+1K)} & ~  
                                        & DeViSE                    & \bf 0.3 & 0.8 & 1.9 & 3.2 & 5.3\\
~                         & {$20,841$}   & \softmaxpp{(1)}          & 0.1 & 1.2 & 3.0 & 4.8 & 7.0\\
~                         & {$+ 1000$}  & \softmaxpp{(10)}          & 0.2 & 1.2 & 3.0 & \bf 5.0 & \bf 7.5\\
~                         & ~           & \softmaxpp{(1000)}        & 0.2 & 1.2 & 3.0 & 4.9 & 7.3 \\
\hline
\end{tabular}
\end{center}
\caption{Flat hit@$k$ performance of DeViSE~\cite{devise} and
  \softmaxpp{\,($T$)} for $T = 1, 10, 1000$ on ImageNet zero-shot
  learning task. When testing the methods with the datasets indicated
  with (+1K), training labels are also included as potential labels
  within the nearest neighbor classifier, hence the number of
  candidate labels is $1000$ more. In all cases, zero-shot classes did
  not occur in the training set, and none of the test images is
  annotated with any of the training labels.}
\label{tab:zeroshot-hit}
\end{table}

\tabref{tab:zeroshot-hit} shows flat hit@$k$ results for the DeViSE
and three versions of the \softmaxpp{} model. The \softmaxpp{} model
has a hyper-parameter $T$ that controls the number of training labels
used for the convex combination of semantic embeddings. We report the
results for $T=1, 10, 1000$ as \softmaxpp{\,($T$)} in
\tabref{tab:zeroshot-hit}. Because there are only $1000$ training
labels, $T$ is bounded by $1 \le T \le 1000$. The results are reported
on the three test datasets; the dataset difficulty increases from top
to bottom in \tabref{tab:zeroshot-hit}. For each dataset, we consider
including and excluding the training labels within the label
candidates used for $k$-nearest neighbor label ranking (\ie~$\Yoset$
in \eqref{eq:zeroshot-extrapolate}). None of the images in the test
set are labeled as training labels, so including training labels in
the label candidate set for ranking hurts the performance as finding
the correct labels is harder in a larger set. Datasets that include
training labels in their label candidate set are marked by
``(+1K)''. The results demonstrate that the \softmaxpp{} model
consistently outperforms the DeViSE on all of the datasets for all
values of $T$. Among different versions of the \softmaxpp{},
\softmaxpp{(10)} performs the best. We do not directly compare against
the method of Socher et al.~\cite{Socher13}, but Frome et
al.~\cite{devise} reported that the ranking loss used within the
DeViSE significantly outperforms the the squared loss used
in~\cite{Socher13}.

Not surprisingly, the performance of the models is best when training
labels are excluded from the label candidate set. All of the models
tend to predict training labels more often than test labels,
especially at their first few predictions. For example, when training
labels are included, the performance of \softmaxpp{(10)} drops from
$9.4\%$ hit@$1$ to $0.3\%$ on the $2$-hops dataset. This suggests that
a procedure better than vanilla $k$-nearest neighbor search needs to
be employed in order to distinguish images that do not belong to the
training labels. We note that the DeViSE has a slightly lower bias
towards training labels as the performance drop after inclusion of
training labels is slightly smaller than the performance drop in the
\softmaxpp{} model.


\tabref{tab:zeroshot-hir} shows hierarchical precision@$k$ results for
the Softmax baseline, DeViSE, and \softmaxpp{(10)} on the zero-shot
learning task. The results are only reported for \softmaxpp{\,(10)}
because $T = 10$ seems to perform the best among $T = 1, 10,
1000$. The hierarchical metric also confirms that the \softmaxpp{}
improves upon the DeViSE for zero-shot learning. We did not compare
against the Softmax baseline on the flat hit@$k$ measure, because the
Softmax model cannot predict any of the test labels. However, using
the hierarchical metric, we can now compare with the Softmax baseline
when the training labels are also included in the label candidate set
(+1K). We find that the top $k$ predictions of the \softmaxpp{}
outperform the Softmax baseline in hierarchical precision@$k$.

\begin{table}[t]
\begin{center}
\begin{tabular}{|l|l|c|c|c|c|c|c|}
\hline
~                                            & ~                    & \multicolumn{5}{c|}{Hierarchical precision@$k$} \\
\cline{3-7}
Test Label Set                               & Model                & 1     & 2     & 5     & 10    & 20    \\
\hline
\multirow{2}{*}{$2$-hops} 
                          & DeViSE                    & 0.06 & 0.152 & 0.192 & 0.217 & 0.233 \\
~                         & \softmaxpp{(10)}          & \bf 0.094 & \bf 0.214 & \bf 0.247 & \bf 0.269 & \bf 0.284 \\
\hdashline
\multirow{3}{*}{$2$-hops (+1K)} 
~                               & Softmax baseline    & 0          & \bf 0.236     & 0.181     & 0.174     & 0.179 \\
~                               & DeViSE              & \bf 0.008  & 0.204     & 0.196     & 0.201     & 0.214 \\
~                               & \softmaxpp{(10)}    & 0.003 & 0.234 & \bf 0.254 & \bf 0.260 & \bf 0.271\\
\hline
\multirow{2}{*}{$3$-hops} 
                          & DeViSE                    & 0.017 & 0.037 & 0.191 & 0.214 & 0.236 \\
~                         & \softmaxpp{(10)}          & \bf 0.027 & \bf 0.053 & \bf 0.202 & \bf 0.224 & \bf 0.247 \\
\hdashline
\multirow{3}{*}{$3$-hops (+1K)} 
~                               & Softmax baseline    & 0 & 0.053 & 0.157 & 0.143 & 0.130 \\
~                               & DeViSE              & \bf 0.005 & 0.053 & 0.192 & 0.201 & 0.214 \\
~                               & \softmaxpp{(10)}    & 0.002 & \bf 0.061 & \bf 0.211 & \bf 0.225 & \bf 0.240 \\
\hline
\multirow{2}{*}{ImageNet 2011 $21$K}       & DeViSE              & 0.008 & 0.017 & 0.072 & 0.085 & 0.096 \\
~                                          & \softmaxpp{(10)}    & \bf 0.014 & \bf 0.025 & \bf 0.078 & \bf 0.092 & \bf 0.104 \\

\hdashline
\multirow{3}{*}{ImageNet 2011 $21$K (+1K)} & Softmax baseline    & 0     & 0.023 & 0.071 & 0.069 & 0.065 \\
~                                          & DeViSE              & \bf 0.003 & 0.025 & 0.083 & 0.092 & 0.101 \\
~                                          & \softmaxpp{(10)}    & 0.002 & \bf 0.029 & \bf 0.086 & \bf 0.097 & \bf 0.105 \\
\hline
\end{tabular}
\end{center}
\caption{Hierarchical precision@$k$ performance of Softmax
  baseline~\cite{Krizhevsky12}, DeViSE~\cite{devise}, and
  \softmaxpp{(10)} on ImageNet zero-shot learning task.}
\label{tab:zeroshot-hir}
\end{table}

\comment{{
Note that the \softmaxpp{(1)} model is simply predicting ...
}}

Even though the \softmaxpp model is proposed for zero-shot learning,
we assess how the \softmaxpp compares with the DeViSE and the Softmax
baseline on the standard classification task with the training $1000$
labels, \ie~the training and test labels are the
same. \tabref{tab:1000:h} and \ref{tab:1000} show the flat hit@$k$ and
hierarchical precision@$k$ rates on the $1000$-class learning
task. According to \tabref{tab:1000:h}, the \softmaxpp{(10)} model
improves upon the Softmax baseline in hierarchical precision at $5$,
$10$, and $20$, suggesting that the mistakes made by the \softmaxpp
model are on average more semantically consistent with the correct
class labels, than the Softmax baseline. This improvement is due to
the use of label embedding vectors learned from Wikipedia
articles. However, on the $1000$-class learning task, the
\softmaxpp{(10)} model underperforms the DeViSE model. We note that
the DeViSE model is trained with respect to a $k$-nearest neighbor
retrieval objective on the same specific set of $1000$ labels, so its
better performance on this task is expected.

Although the DeViSE model performs better than the \softmaxpp on the
original $1000$-class learning task (\tabref{tab:1000:h},
\ref{tab:1000}), it does not generalize as well as the \softmaxpp
model to the unseen zero-shot learning categories
(\tabref{tab:zeroshot-hit}, \ref{tab:zeroshot-hir}). Based on this
observation, we conclude that a better $k$-nearest neighbor
classification on the training labels, does not automatically
translate into a better $k$-nearest neighbor classification on a
zero-shot learning task. We believe that the DeViSE model suffers from
a variant of overfitting, which is the model has learned a highly
non-linear and complex embedding function for images. This complex
embedding function is well suited for predicting the training label
embeddings, but it does not generalize well to novel unseen label
embedding vectors. In contrast, a simpler embedding model based on
convex combination of semantic embeddings (\softmaxpp) generalizes
more reliably to unseen zero-shot classes, with little chance of
overfitting.

\begin{table}[t]
\begin{center}
\begin{tabular}{|l|l|c|c|c|c|c|c|}
\hline
~                                            & ~                    & \multicolumn{5}{c|}{Hierarchical precision@$k$} \\
\cline{3-7}
Test Label Set                               & Model                & 1     & 2     & 5     & 10    & 20    \\
\hline
\multirow{5}{*}{ImageNet 2011 $1$K} & Softmax baseline    & \bf 0.556    & \bf 0.452 & 0.342 & 0.313 & 0.319 \\
 & DeViSE & 0.532 & 0.447 & \bf 0.352 & \bf 0.331 & \bf 0.341 \\
 & \softmaxpp(1) & 0.551 & 0.422 & 0.32 & 0.297 & 0.313 \\
 & \softmaxpp(10) & 0.543 & 0.447 & 0.348 & 0.322 & 0.337 \\
 & \softmaxpp(1000) & 0.539 & 0.442 & 0.344 & 0.319 & 0.335 \\
\hline
\end{tabular}
\end{center}
\caption{Hierarchical precision@$k$ performance of Softmax
  baseline~\cite{Krizhevsky12}, DeViSE~\cite{devise}, and
  \softmaxpp on ImageNet original 1000-class learning task.}
\label{tab:1000:h}
\end{table}

\begin{table}[t]
\begin{center}
\begin{tabular}{|l|l|c|c|c|c|c|}
\hline
~                                            & ~                    & \multicolumn{5}{c|}{Flat hit@$k$ (\%)} \\
\cline{3-6}
Test Label Set                               & Model                & 1     & 2     & 5     & 10     \\
\hline
\multirow{5}{*}{ImageNet 2011 $1$K} & Softmax baseline    & \bf 55.6     & \bf 67.4 & \bf 78.5 & \bf 85.0 \\
 & DeViSE & 53.2 & 65.2 & 76.7 & 83.3 \\
 & \softmaxpp(1) & 55.1 & 57.7 & 60.9 & 63.5 \\
 & \softmaxpp(10) & 54.3 & 61.9 & 68.0 & 71.6 \\
 & \softmaxpp(1000) & 53.9 & 61.1 & 67.0 & 70.6 \\
\hline
\end{tabular}
\end{center}
\caption{Flat hit@$k$ performance of Softmax
  baseline~\cite{Krizhevsky12}, DeViSE~\cite{devise}, and
  \softmaxpp on ImageNet original 1000-class learning task.}
\label{tab:1000}
\end{table}

{\bf Implementation details.} The \softmaxpp{(1)} model takes the
top-$1$ prediction of the convolutional net, and expands it to a list
of labels based on the similarity of the label embedding
vectors. To implement \softmaxpp{(1)} efficiently, one can
pre-compute a list of test labels for each training label, and simply
predict the corresponding list based on the top prediction of the
convolutional net. The top prediction of the \softmaxpp{(1)}
occasionally differs from the top prediction of the Softmax baseline
due to a detail of our implementation. In the Imagenet experiments,
following the setup of the DeViSE model, there is not a one-to-one
correspondence between the class labels and the word embedding
vectors. Rather, because of the way the Imagenet synsets are defined,
each class label is associated with several synonym terms, and hence
several word embedding vectors. In the process of mapping the Softmax
scores to an embedding vector, the ConSE model first averages the word
vectors associated with each class label, and then linearly combine
the average vectors according to the Softmax scores. However, when we
rank the word vectors to find the k most likely class labels, we
search over individual word vectors, without any averaging of the
synonym words. Thus, the ConSE(1) might produce an average embedding
which is not the closest vector to any of the word vectors
corresponding to the original class label, and this results in a
slight difference in the hit@1 scores for ConSE(1) and the Softmax
baseline. While other alternatives exist for this part of the
algorithm, we intentionally kept the ranking procedure exactly the
same as the DeViSE model to perform a direct comparison.

%% file: conclusion.tex
\comment{{
\section{Conclusion}

In spite of the simplicity of the
\softmaxpp{(1)} and \softmaxpp{(10)} models, they outperform the
DeViSE model based on hit@$k$ for $k = 2, 5, 10, 20$ on all of the
three datasets. 
\tabref{tab:zeroshot-hit} shows that the \softmaxpp{} model
consistently outperforms the DeViSE on both flat and hierarchical
metrics.
}}

\section{Conclusion}
\label{section:conclusion}

The \softmaxpp approach to mapping images into a semantic embedding
space is deceptively simple.  Treating classifier scores as weights in
a convex combination of word vectors is perhaps the most direct method
imaginable for recasting an \nway image classification system as image
embedding system.  Yet this method outperforms more elaborate joint
training approaches both on zero-short learning and on performance
metrics which weight errors based on semantic quality.  The success of
this method undoubtedly lays is its ability to leverage the strengths
inherent in the state-of-the-art image classifier and the
state-of-the-art text embedding system from which it was constructed.

While it draws from their strengths, we have no reason to believe that
\softmaxpp depends on the details the visual and text models from
which it is constructed.  In particular, though we used a deep
convolutional network with a Softmax classifier to generate the
weights for our linear combination, any visual object classification
system which produces relative scores over a set of classes is
compatible with the \softmaxpp framework.  Similarly, though we used
semantic embedding vectors which were the side product of an
unsupervised natural language processing task, the \softmaxpp
framework is applicable to other alternative representation of text in
which similar concepts are nearby in vector space. The choice of the
training corpus for the word embeddings affects the results too.

One feature of the \softmaxpp model which we did not exploit in our
experiments is its natural representation of confidence.  The norm of
the vector that \softmaxpp assigns to an image is a implicit expression of
the model's confidence in the embedding of that image.  Label
assignments about which the Softmax classifier is uncertain be given
lower scores, which naturally reduces the magnitude of the \softmaxpp
linear combination, particularly if Softmax probabilities are used as
weights without renormalization.  Moreover, linear combinations of
labels with disparate semantics under the text model will have a lower
magnitude than linear combinations of the same number of closely
related labels.  These two effects combine such that \softmaxpp only
produces embeddings with an L2-norm near 1.0 for images which were
either nearly completely unambiguous under the image model or which
were assigned a small number of nearly synonymous text labels.  We
believe that this property could be fruitfully exploited in settings
where confidence is a useful signal.